\pdfoutput=1

\documentclass[11pt]{article}

\usepackage{ACL2023}

\usepackage{times}
\usepackage{latexsym}
\usepackage{booktabs}
\usepackage{graphicx}
\usepackage[T1]{fontenc}

\usepackage[utf8]{inputenc}

\usepackage{microtype}

\usepackage{inconsolata}
\usepackage{multirow}
\usepackage{makecell}
\usepackage[linesnumbered,ruled,vlined]{algorithm2e}
\usepackage{color}
\usepackage{amsmath}

\title{Communication Efficient Federated Learning for Multilingual \\ Neural Machine Translation with Adapter}

\author{
Yi Liu\textsuperscript{1},
Xiaohan Bi\textsuperscript{2},
Lei Li\textsuperscript{1}, Sishuo Chen\textsuperscript{2}, Wenkai Yang\textsuperscript{2}, Xu Sun\textsuperscript{1}\\ 
      \textsuperscript{1}National Key Laboratory for Multimedia Information Processing,\\
      School of Computer Science, Peking University \\ 
      \textsuperscript{2}Center for Data Science, Peking University
\\ \texttt{yliu.pku@outlook.com}\quad \texttt{\{bxh,wkyang\}@stu.pku.edu.cn}
\\ \texttt{nlp.lilei@gmail.com}\quad \texttt{\{chensishuo,xusun\}@pku.edu.cn}  
  }

\begin{document}
\maketitle
\begin{abstract}

Federated Multilingual Neural Machine Translation (Fed-MNMT) has emerged as a promising paradigm for institutions with limited language resources. This approach allows multiple institutions to act as clients and train a unified model through model synchronization, rather than collecting sensitive data for centralized training. This significantly reduces the cost of corpus collection and preserves data privacy. However, as pre-trained language models~(PLMs) continue to increase in size, the communication cost for transmitting parameters during synchronization has become a training speed bottleneck.
In this paper, we propose a communication-efficient Fed-MNMT framework that addresses this issue by keeping PLMs frozen and only transferring lightweight adapter modules between clients. Since different language pairs exhibit substantial discrepancies in data distributions, adapter parameters of clients may conflict with each other. To tackle this, we explore various clustering strategies to group parameters for integration and mitigate the negative effects of conflicting parameters.
Experimental results demonstrate that our framework reduces communication cost by over 98\% while achieving similar or even better performance compared to competitive baselines. Further analysis reveals that clustering strategies effectively solve the problem of linguistic discrepancy and pruning adapter modules further improves communication efficiency.\footnote{Our code is available at \url{https://github.com/lancopku/FedMNMT}}
\end{abstract}

\section{Introduction}
Federated Learning (FL)~\citep{mcmahan2017communication} provides a new training framework utilizing data from various clients without privacy leakage. In FL, the server receives models from clients trained with their local data and aggregates all parameters it has received to acquire a global model, and then sends it back to all clients to start the next training round.
This characteristic enables FL to get widely applied in real-world scenarios~\citep{ge2020fedner,roosta2021communication,passban2022training,niu2022federated}. 
In recent years, federated multilingual neural machine translation (Fed-MNMT) has become a new training paradigm and making it feasible for most institutions to train MNMT models \cite{roosta2021communication,passban2022training}. FL makes it possible to leverage corpora from other organizations without privacy problems, solving the problem that training an MNMT model needs to collect large-scale multilingual corpora, which is expensive, time-consuming, and often unaffordable for resource-constrained institutions. 
Therefore, Fed-MNMT is a secure and cost-effective alternative to conventional centralized training for the optimization of MNMT models. \par

\begin{figure}[t] \centering
    \includegraphics[width=\linewidth]{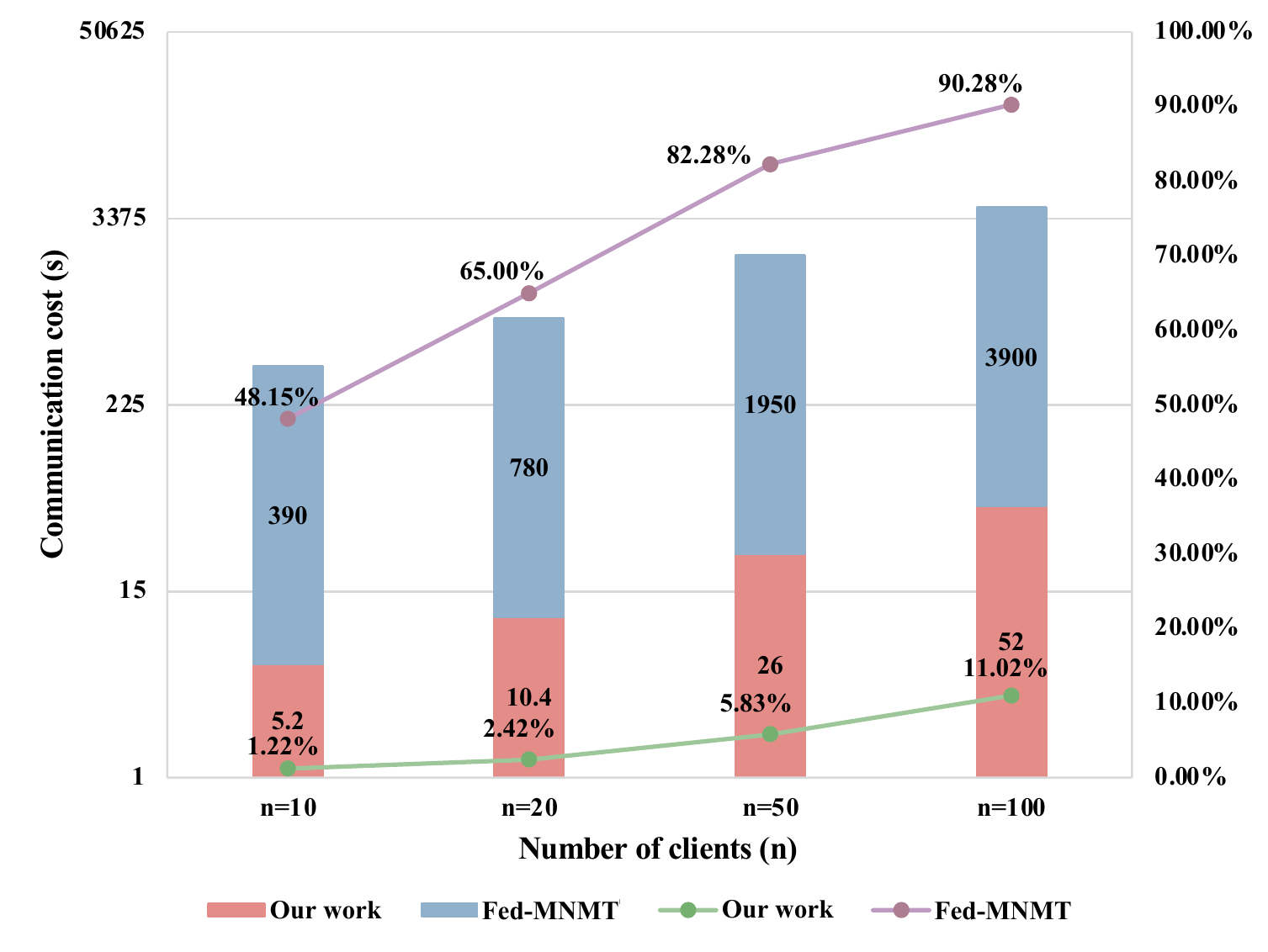}
    \caption{Estimated transfer time (bar chart) and the approximate ratio of transfer time to total training time (line chart) in one round for a low-resource client. Our method significantly reduces communication cost and improve training efficiency.}
    \label{fig:comm_cost} 
\end{figure}

\begin{figure*}[ht] \centering
    \includegraphics[width=0.85\linewidth]{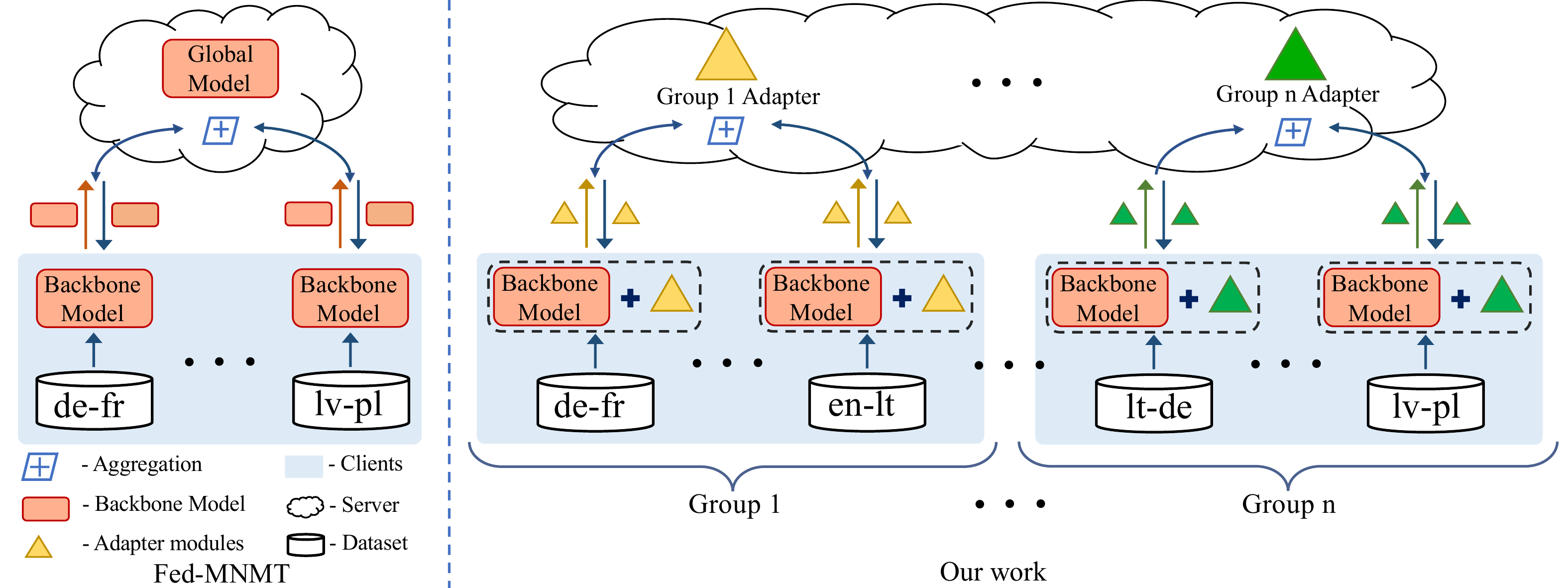}
    \caption{Communication-efficient Fed-MNMT framework with adapter modules and clients clustering strategies.}
    \label{fig:intro} 
\end{figure*}

However, the issue of communication cost is non-negligible when we introduce FL to neural machine translation. 
Unlike local centralized learning, federated learning requires frequent communication of model parameters between the server and clients. 
Therefore, the communication cost grows rapidly along with the increase in model size. 
Nowadays, pre-trained language models are widely adopted as backbone models for MNMT, whose parameters are usually over $10^8$, e.g., 611M for mBART-50~\citep{Tang2020MultilingualTW} and 1.2B for M2M100~\cite{fan2020englishcentric}.
Considering the increasing number of clients in realistic scenarios as there frequently appear new clients, communication costs will severely hinder the efficient training of the entire Fed-MNMT system and thus make the application of FL to MNMT impractical.

To tackle this problem, we introduce the parameter-efficient tuning idea~\citep{houlsby2019parameter, pfeiffer2020adapterfusion, karimi2021compacter} into Fed-MNMT. 
Specifically, we focus on adapter~\citep{rebuffi2017learning,houlsby2019parameter}, which is a popular technique for efficient tuning that requires only updating a lightweight adapter module.
In the training process of adapters, a number of randomly initialized modules are inserted into the backbone models and fine-tuned on new data.
Concretely, only the parameters of these modules are updated during training, so the number of parameters needed to be transferred between the server and clients is substantially reduced.  
As the communication cost before and after introducing adapter illustrated in Figure~\ref{fig:comm_cost}, this approach significantly saves communication costs and enables practical applications of Fed-MNMT. 

However, directly adding adapter modules to NMT models results in a performance decline, which is initially observed by \citet{roosta2021communication} and also confirmed by our experimental results.
This phenomenon is attributed to the divergence of different language pairs. In Fed-MNMT, corpora in diverse languages from different clients are not independently and identically distributed (Non-I.I.D.), so directly aggregating parameters from clients leads to a decrease in model's performance~\citep{zhao2018federated}.\par
Considering the adverse effect of conflicting parameters from diverse languages in Fed-MNMT, we introduce clustering strategies to alleviate this issue. The core idea is to cluster the samples according to the characteristics of their data and only conduct aggregation within each cluster where samples share similar properties. 
Specifically, we cluster all clients with different language pairs based on the language family, gradient similarity, and random, respectively, and systematically compare the performance of different clustering strategies on multilingual translation benchmarks.
Figure~\ref{fig:intro} gives a general view of our training framework.
Our experimental results show that clustering on adapters alleviates the data Non-I.I.D.\ problem and yields better performance in most cases. 
Overall, our work opens a new direction for future improvements on Fed-MNMT in the real world. \par
In conclusion, our primary contributions can be summarized as follows:
\begin{itemize}
    \item Aware of the communication barrier in the training of Fed-MNMT models, we introduce a practical efficient Fed-MNMT framework to enable real-world applications.
    \item By exploring adapter and clustering strategies for alleviating the undesirable effect of data discrepancy, we achieve comparable results with over 98\% communication cost reduced compared to vanilla Fed-MNMT.
\end{itemize}

\section{Methodology}
In this section, we first define the Fed-MNMT problem in \S~\ref{subsec:formulation}. 
Next, we elaborate on the adapter modules and the investigated clustering strategies in \S~\ref{subsec:adapterModuels} and \S~\ref{subsec:clustering}, respectively. 
Last, we provide an analysis of communication costs between the original Fed-MNMT and our method in \S~\ref{subsec:comparison}.
\subsection{Problem Formulation} \label{subsec:formulation}
For a Fed-MNMT problem, we suppose that the set of clients is $\left\{C_i\right\}_{i=1}^N$ where $N>1$ and client $C_i$ owns only one language pair $P_i$, whose source language and target language are $src_i$ and $tgt_i$, respectively, and corresponding dataset $D_i=\{x_{ij},y_{ij}\}_{j=1}^{n_i}$, where $n_i$ is the size of $D_i$. In each training round, the optimization target for $C_i$ is minimizing the cross entropy loss between the ground truths $y_i$ and model's predictions $\tilde{y}_i$:
\begin{equation}
\mathcal{L}_i = -\sum_{j=1}^{n_i}\sum_{k=1}^{l_{ij}}\log p\left(\tilde{y}_{ij}^k=y_{ij}^k|x_{ij}\right)
\end{equation}
After the $t$-th training round, all clients will deliver local parameters to the server. The server will aggregate these parameters to obtain the initial parameters for the next round's training. A commonly adopted aggregation algorithm is FedAvg~\citep{mcmahan2017communication}, where the weighted average of clients' parameters is calculated according to the quantities of local data samples. Let $\Theta$ denote model parameters, the FedAvg algorithm can be formulated as: 
\begin{equation}
\label{eq:fedavg}
\Theta^{t+1}=\sum_{i=1}^N \frac{n_i}{n}\Theta_i^t,
\end{equation}
where $n=\sum_{i=1}^N n_i$. Then the aggregated parameters will be sent back to all clients to initialize their local models for the next round of training.\par
However, data sizes can vary sharply among low-resource and high-resource languages in Fed-MNMT and FedAvg cannot deal well with data quantity skew well~\citep{wang2020tackling}. Thus we change FedAvg, calculating the weighted mean of different clients' parameters, to directly calculating the arithmetic mean of parameters:
\begin{equation}
\label{eq:fedmean}
\Theta^{t+1}=\sum_{i=1}^N \frac{1}{N}\Theta_i^t
\end{equation}
We refer to this aggregation method as FedMean in our paper.\par
Considering the size of pre-trained multi-language models, the communication of model parameters between the server and clients is time-consuming. Inspired by recent progress in parameter efficient tuning, we are interested in whether adapter can be used to improve the efficiency in FL.\looseness=-1

\subsection{Adapter Modules} \label{subsec:adapterModuels}
We introduce bottleneck adapter~\citep{houlsby2019parameter} into pre-trained multilingual models. Following the settings of \citet{houlsby2019parameter} and \citet{pfeiffer2020AdapterHub}, we add adapter modules after the self-attention layer and  feed-forward network (FFN) layer for each encoder layer and an additional adapter layer after the cross-attention layer for each decoder layer. During training, only the parameters of adapters and layer-norm modules will be updated thus only a small proportion of parameters have to be communicated between the server and clients. \par

\subsection{Client Clustering Strategies}
\label{subsec:clustering}
Related research~\citep{johnson2017google,firat2016multi} has shown that parameter sharing among different languages in MNMT boosts the model's performance, especially for those low-resource languages. 
Motivated by the success of language clustering in MNMT~\citep{tan2019multilingual}, we decide to introduce the method of language pairs clustering into the Fed-MNMT problem and we only allow inner-cluster parameters aggregation. Assuming that the multi-language model consists of an encoder and a decoder, we first conduct a clustering algorithm to obtain the encoder clusters set $G_e=\{g_i\}_{i=1}^{m_e}$ and the decoder clusters set $G_d=\{g_i\}_{i=1}^{m_d}$. Each cluster $g_i$ contains the ids of clients in this cluster. Detailed aggregation algorithm is shown in Algorithm \ref{alg:cluster}.  \par

\begin{algorithm}[ht]
\small
\caption{Inner-cluster Aggregation}
\label{alg:cluster}
\KwIn{Encoder and decoder clusters sets $G_e$ and $G_d$\;\hspace{3em}Initial encoder and decoder paras $\Theta_e^0$ and $\Theta_d^0$\;\hspace{3em}Clients set $\{C_i\}_{i=1}^N$\;\hspace{3em}Training round $T$.}
\KwOut{Encoder paras $\{\Theta_{e,i}^T\}_{i=1}^N$\;\hspace{3.75em}Decoder paras $\{\Theta_{d,i}^T\}_{i=1}^N$}.
\For{$i$ from $1$ to $N$}{
    Initialize $\Theta_{e,i}^0$ with $\Theta_e^0$\;
    Initialize $\Theta_{d,i}^0$ with $\Theta_d^0$\;
}
\For{$t$ from $1$ to $T$}{
    \For{$i$ from $1$ to $N$}{
        \tcp{local update of client $i$}
        update $\Theta_{e,i}^{t-1}$ and $\Theta_{d,i}^{t-1}$ with local data\;
    }
    \tcp{inner-cluster aggregation of encoder parameters} 
    \ForEach{$g$ in $G_e$}{
        $\Theta_{e,g}^t=\sum_{id \in g} \frac{1}{\vert g \vert}\Theta_{e,id}^{t-1}$\;
        \ForEach{$id$ in $g$}{
            $\Theta_{e,id}^t = \Theta_{e,g}^t$\;
        }
    }
    \tcp{inner-cluster aggregation of decoder parameters}
    \ForEach{$g$ in $G_d$}{
        $\Theta_{d,g}^t=\sum_{id \in g} \frac{1}{\vert g \vert}\Theta_{d,id}^{t-1}$\;
        \ForEach{$id$ in $g$}{
            $\Theta_{d,id}^t = \Theta_{d,g}^t$\;
        }
    }
}
\end{algorithm}

We explore the following three different clustering strategies. \par
\paragraph{Language families/groups.} \citet{chronopoulou2022language} have verified the strategy of sharing parameters within the same language family in the MNMT problem. We decide to use this strategy in the FL setting. We choose 8 languages belonging to 4 different language families from the TED2020 corpus and 10 languages belonging to 4 different language groups, which are all parts of the Indo-European language family, from the Europarl corpus. The clustering of the encoder depends on language families/groups of source languages and the clustering of the decoder is decided by the target languages' families/groups. Languages from the same family or group will be clustered into the same group. \par 
\paragraph{Gradients.} Unlike in the scene of centralized learning, clustering based on model parameters~\citep{tan2019multilingual} in Fed-MNMT is unfeasible due to privacy problems. Therefore, we use gradients as the basis of the feature for clustering instead. For each language pair, we use a pre-trained multi-language model to acquire an average gradient vector of all data samples, then a clustering algorithm is applied to the gradient vectors in order to separate clients into different groups. The number of parameters we use for gradients clustering is only about 131K for each client and will hardly introduce any extra communication cost. \par
\paragraph{Random clustering.} We also test randomly separating all clients into different groups  as a baseline for clustering strategies. In detail, we uniformly separate the clients and keep the numbers of clusters in the encoder and the decoder the same as those in the language families/groups strategy.\par

\subsection{Communication Cost Comparison}\label{subsec:comparison}
Taking the mBART-50 model~\citep{Tang2020MultilingualTW}, which is a popular pre-trained multi-language model, as an example, the number of parameters is around 610.9M, which requires about 2.44GB storage space in the FP32 format. In comparison, after adding adapter modules, only about 8M parameters have to be transferred, which will save approximately $98.7$\% communication cost.\par
More concretely, we provide an approximation for the transmission time needed between the server and clients as follows. Assuming the maximum bandwidth of the server is $1000$Mbps, the time to transfer the entire mBART model from client to server is around $2.44$GB / $1000$Mbps = $19.5$ seconds. Assuming all clients share the bandwidth, the total transfer time grows linearly with the number of clients. The synchronization process for all clients to finish transferring models to the server will occupy a large proportion of the total training time. In our actual experiments with $12 $ clients, the theoretical total transferring time is about $19.5 \times 12 = 234$ seconds. However, for clients with low-resource languages, the training could be finished within $7$ minutes, which means transferring time occupies over half of the local training time. By contrast, the time to transfer the adapter's parameters is only about $0.26$ seconds, which is negligible compared to local training time thus significantly improving the training efficiency. \par 

\section{Experimental Setup}
\subsection{Datasets and Evaluation Metrics}

We conduct experiments in two different settings: Multi-to-English and Multi-to-Multi (hereinafter referred to as ``m2en'' and ``m2m'', respectively). We use the \textbf{TED2020} corpus~\citep{reimers-2020-multilingual-sentence-bert} for the m2en setting and the \textbf{Europarl} corpus~\citep{koehn2005europarl} for the m2m setting. The TED2020 corpus is extracted from TED speeches and contains over 100 languages around the world. The Europarl corpus is from the proceedings of the European Parliament and contains 21 languages of European countries. \par
For each language pair\footnote{Abbreviations for languages we use: Chinese->zh, English->en, Thai->th, Arabic->ar, Hebrew->he, Finnish->fi, Estonian->et, Russian->ru, Slovene->sl, German->de, Dutch->nl, French->fr, Italian->it, Spain->es, Polish->pl, Slovene->sl, Lithuanian->lt, Latvian->lv.}, we divide the original corpus into training, dev, and test datasets in accordance of the proportion of 6:2:2. We further sample subsets from the divided datasets. To simulate the scene of low-resource languages and high-resource languages, the training data size of each language pair varies according to the corresponding original corpus size. The specific language pairs we use and corresponding data sizes are shown in Appendix \ref{appendix:data_size}.\looseness=-1

In the m2en setting, for clustering strategies based on language families/groups and random shuffle,  clustering algorithms are only applied to the encoder and all clients share the decoder's parameters because their target languages are all English. But for clustering based on gradients, we also cluster the decoder's parameters into different groups and the number of groups stays the same as that in the encoder. In the m2m setting, clustering is conducted for both the encoder and the decoder. Meanwhile, the numbers of groups in the encoder and decoder are the same in all clustering strategies. We will provide a further analysis of the clustering strategies of the m2m setting in \S~\ref{subsec:ablation_study}.\looseness=-1

We choose the BLEU score as the evaluation metric using the SacreBLEU~\citep{post-2018-call} package. Aside from the BLEU score on each language pair, we additionally report the macro average and micro average scores on all language pairs.

\subsection{Baselines}
We evaluate the following methods as baselines:\par
\textbf{Centralized-model}. The results of centralized training, where data from all clients are gathered together, using the original multi-language model without extra modules. \par
\textbf{Centralized-adapter}. The results of centralized training using the multi-language model with adapter modules. \par
\textbf{Adapter-local}. We train a model for each client using local data without parameter aggregation with other language pairs. \par
\textbf{Model-fed}. We train the original multi-language model without Adapter modules under the federated learning framework, where the parameters are shared among all clients using the aggregation algorithm in Eq.~\eqref{eq:fedmean} without any clustering strategies. \par
\textbf{Adapter-fed}. In this method, adapter modules are attached to the backbone model, while the rest settings are the same as those in \emph{model-fed}. This baseline corresponds to the scene of directly introducing adapter without any clustering strategies.\looseness=-1

\subsection{Training Setup}
We choose the mBART-50 pre-trained model \footnote{\url{https://huggingface.co/facebook/mbart-large-50-many-to-many-mmt}} as our backbone model.
To fairly compare the training and communication costs of different methods, we train each model for 5 rounds. We select the checkpoint with the lowest loss on the dev set and evaluate it on the test set. Parameters are aggregated every time all clients finish an epoch of local training. For every client, the batch size is 8 and the local model is updated every 16 steps. The local learning rate is  $5\times10^{-5}$ for the mBART model and $1\times10^{-3}$ for the models with adapter modules. The hidden size of adapter modules is 64.\par
For all experiments, we train the model with 3 random seeds and report the average scores. For the random clustering strategy, the clustering groups are different when using different random seeds. 
\begin{table*}[h]
    \centering
    \resizebox{0.95\textwidth}{!}{
    \begin{tabular}{@{}l|c|cccccccc|cc@{}}
    \toprule
   Method  & Comm. Cost & zh-en & th-en & ar-en & he-en & fi-en & et-en & ru-en & sl-en & Macro Avg. & Micro Avg. \\
    \midrule
    centralized-model & N / A & 24.72 & 28.97 & 38.29 & 43.59 & 32.81 & 32.70 & 30.14 & 47.92 & 34.89 & 32.33 \\
    centralized-adapter & N / A & 24.97 & 21.47 & 39.02 & 45.05 & 33.62 & 32.62 & 30.65 & 50.45 & 34.73 & 32.03 \\
    adapter-local & N / A & 25.16 & 21.68 & 39.46 & 44.32 & 33.12 & 32.64 & 30.58 & 50.93 & 34.73 & 32.15 \\
    model-fed & 611M & 25.31 & 23.41 & 39.54 & 45.13 & 32.87 & 33.27 & 30.77 & 51.85 & 35.27 & 32.55 \\
    \midrule
    adapter-fed & 8M & 25.12 & 16.64 & \textbf{39.62} & \textbf{44.93} & 33.14 & 32.66 & 30.41 & \textbf{53.62} & 34.52 & 31.71 \\
    adapter-random & 8M & \textbf{25.37} & 21.48 & 39.61 & 44.82 & 33.26 & 33.21 & 30.75 & 52.64 & 35.14 & 32.38 \\
    adapter-gradients & 8M & 25.26 & 21.26 & 39.39 & 44.64 & 33.62 & 33.14 & 30.72 & 51.16 & 34.90 & 32.21 \\
    adapter-families & 8M & 25.28 & \textbf{21.58} & 39.45 & 44.70 & \textbf{33.87} & \textbf{33.23} & \textbf{30.92} & 52.64 & \textbf{35.21} & \textbf{32.40} \\
    \bottomrule
    \end{tabular}}
    \caption{BLEU scores on the TED2020 corpus. Comm. Cost, which is short for communication cost, denotes the number of parameters communicated between the server and each client. \emph{Adapter-random}, \emph{adapter-gradients}, and \emph{adapter-families} refer to clustering strategies of random clustering, gradients, and language families/groups, respectively. The best result of each language pair is highlighted in \textbf{bold} (only methods with adapter modules trained in the FL setting are considered).} 
    \label{tab:ted_results}
\end{table*}

\begin{table*}[h]
    \centering
    \resizebox{0.95\textwidth}{!}{ 
    \begin{tabular}{@{}l|c|cccccccccccc|cc@{}}
    
    \toprule
    Method & Comm. Cost & de-fr & nl-pl & en-lt & fr-nl & it-sl & es-lv & pl-en  & sl-es & sl-lt & lt-de & lv-it & lv-pl & Macro Avg. & Micro Avg.  \\
    \midrule
    centralized-model & N / A & 30.43 & 19.15 & 28.86 & 23.20 & 19.96 & 27.17 & 40.35 & 33.20 & 21.24 & 21.84 & 23.63 & 20.57 & 25.80 & 26.10 \\
    centralized-adapter & N / A & 30.59 & 19.07 & 29.65 & 22.92 & 18.68 & 27.73 & 41.52 & 33.20 & 21.45 & 22.16 & 23.40 & 20.85 & 25.93 & 26.19 \\
    adapter-local & N / A & 30.88 & 19.19 & 30.31 & 23.50 & 20.01 & 28.12 & 41.84 & 33.39 & 21.13 & 22.27 & 23.62 & 21.05 & 26.28 & 26.56 \\
    model-fed & 611M & 30.41 & 17.60 & 29.62 & 19.76 & 13.41 & 28.01 & 39.77 & 32.25 & 21.10 & 21.94 & 20.09 & 19.92 & 24.49 & 24.68 \\
    \midrule
    adapter-fed & 8M & 29.75 & 17.47 & 30.08 & 16.92 & 11.85 & 28.01 & 38.18 & 31.06 & 20.18 & 21.23 & 18.02 & 19.97 & 23.56 & 23.53 \\
    adapter-random & 8M & \textbf{30.90} & 18.57 & 30.04 & 22.39 & 16.53 & \textbf{28.25} & 36.97 & 33.04 & 21.13 & \textbf{22.28} & \textbf{22.72} & 20.37 & 25.27 & 25.67 \\
    adapter-gradients & 8M & 30.14 & \textbf{19.69} & \textbf{30.19} & 19.53 & \textbf{19.14} & 28.06 & \textbf{41.73} & \textbf{33.31} & \textbf{21.51} & 21.60 & 19.95 & \textbf{21.29} & 25.51 & 25.35 \\
    adapter-families & 8M & 30.60 & 19.31 & 30.12 & \textbf{22.65} & 16.69 & 27.99 & 41.33 & 33.21 & 21.31 & 22.04 & 22.68 & 21.21 & \textbf{25.76} & \textbf{26.04} \\
    \bottomrule
    \end{tabular}}
    \caption{BLEU scores on the Europarl corpus. The meanings of symbols stay the same as those in Table \ref{tab:ted_results}.} 
    \label{tab:ep_results}
\end{table*}

\section{Experimental Results}
\label{section:main_results}
\subsection{Primary Results and Findings}
\label{subsec:primary_findings}
The experimental results in the m2en and m2m settings are shown in Table~\ref{tab:ted_results} and Table~\ref{tab:ep_results}, respectively.
In general, directly adding adapter modules leads to a performance drop (comparing \emph{adapter-fed} and \emph{model-fed}) and methods with clustering strategies all achieve better performance than the direct baseline \emph{adapter-fed}, indicating the ability of our clustering strategies to alleviate data discrepancy. In both settings, \emph{adapter-families} method performs best in macro and micro average scores among three clustering strategies, even surpassing \emph{model-fed} in the m2m setting. 

It is noteworthy that the clustering strategies acquire more significant performance improvements in the m2m setting than in the m2en setting. The problem of conflicting parameters is more nettlesome in the m2m setting because there exist more kinds of languages (especially target languages). Thus introducing clustering strategies will bring more benefit to m2m translation tasks.

Meanwhile, we notice that our clustering strategies fail to beat \emph{adapter-local} in the m2m setting. This can also be explained by the difference in the difficulty of tasks. In the more complicated m2m translation task, more elaborate clustering strategies should be designed to fully make advantage of other language pairs and avoid the influence of conflicting parameters. However, we bring the ability of multi-language translation to these clients through FL with an acceptable drop in performance compared to \emph{adapter-local}.

\subsection{Ablation Study}
\label{subsec:ablation_study}
\begin{figure}[t] \centering
    \includegraphics[width=\linewidth]{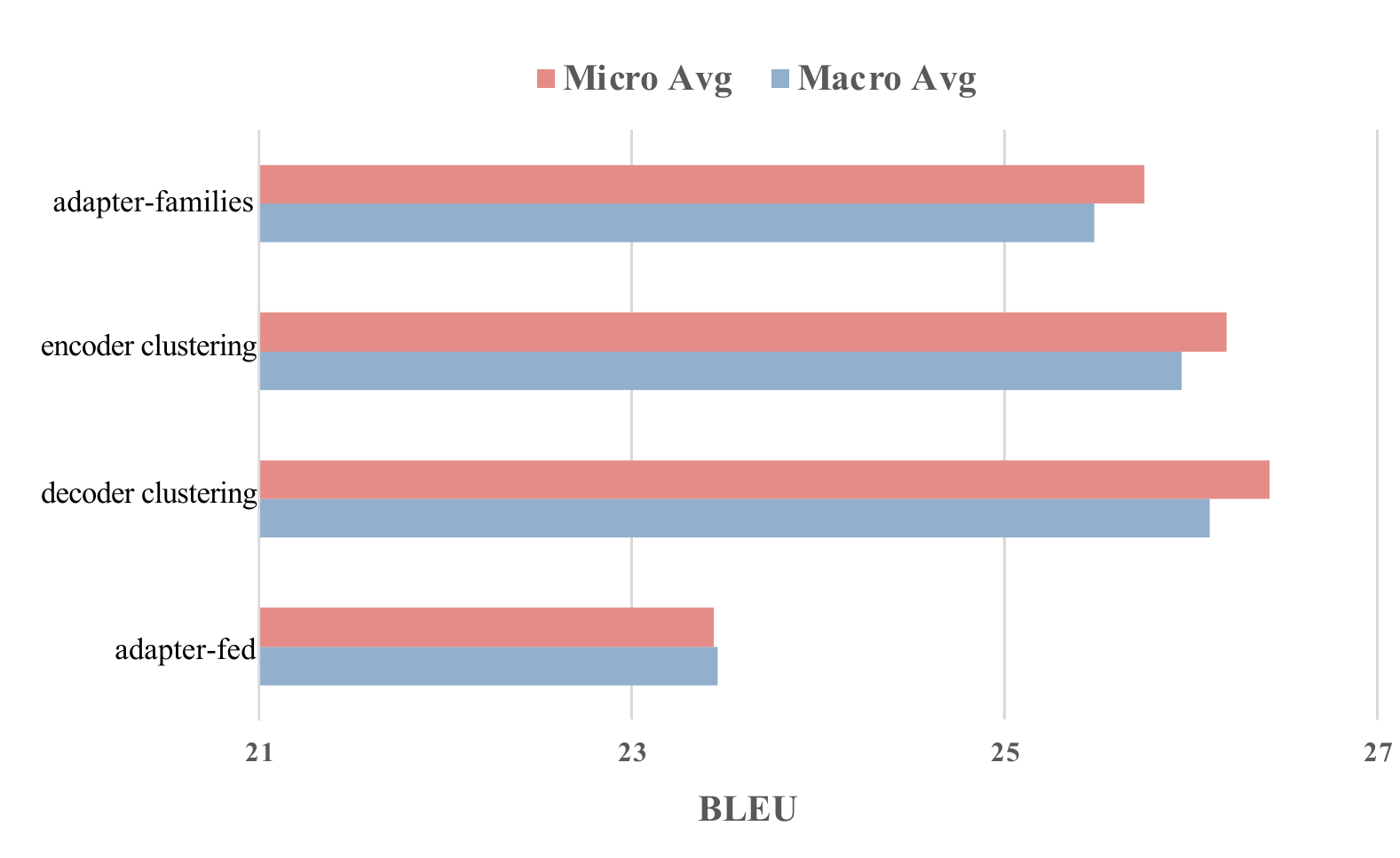}
    \caption{Results of ablation study. Encoder/Decoder-cluster denote clustering only in the encoder/decoder, both obtaining significant improvements compared to the no-clustering baseline.}
    \label{fig:ablation_study} 
\end{figure}

In the m2m setting, the clustering of the adapter modules attached to the encoder and the decoder are independent.
To further explore the specific influence of clustering in these two modules on the model's performance, we apply the clustering strategy to only the encoder and the decoder separately and show the results in Figure~\ref{fig:ablation_study}. We use language families/groups as the clustering strategy. All methods are trained in the m2m setting using the Europarl dataset with one random seed. Other settings are the same as those in our main experiments.\looseness=-1

To our surprise, either clustering in the encoder or the decoder significantly improves performance compared to no-sharing strategies, even surpasses \emph{adapter-families}. We owe this phenomenon to our naive clustering strategies in the encoder and the decoder. In \emph{adapter-families}, the clustering of the encoder and the decoder is only related to the source and target languages respectively. However, parameters in both the encoder and the decoder are influenced by source and target languages together during training. 
The inconsistency between clustering strategies and parameter update results in  \emph{adapter-families} method's worse performance than \emph{adapter-encoder} and \emph{adapter-decoder}.

\begin{table}[ht]
    \centering
    \resizebox{0.95\linewidth}{!}{
    \begin{tabular}{@{}l|l@{}}
    \toprule
    Method & Sentence \\
    \midrule
    Ground truth & I think if somebody tells a lie, \textcolor{blue}{they're not just a liar}. \\
    adapter-families & I think if somebody tells a lie, \textcolor{blue}{they're not just a liar}. \\
    adapter-fed & I know that when people lie, \textcolor{red}{they're just lying}. \\
    \midrule
    Ground truth & I \textcolor{blue}{never miss} a single training session.\\
    adapter-families & I \textcolor{blue}{didn't miss} a single training session.\\
    adapter-fed & I \textcolor{red}{missed} one training session.\\
    \midrule
    \midrule
    Ground truth & And within \textcolor{blue}{four or five weeks}, he can do it again. \\
    adapter-families & And within \textcolor{blue}{four or five weeks}, he can do it again. \\
    adapter-fed & And within \textcolor{red}{a day or two}, he can do this. \\
    \midrule
    Ground truth & So could art \textcolor{blue}{change} the world? \\
    adapter-families & Can art \textcolor{blue}{change} the world? \\
    adapter-fed & Is art about \textcolor{red}{saving} the world? \\
    \midrule
    \midrule
    Ground truth & Over 100,000 \textcolor{blue}{children} learn \textcolor{blue}{science} this way. \\
    adapter-families & Hundreds of thousands of \textcolor{blue}{children} learn \textcolor{blue}{science} this way. \\
    adapter-fed & Hundreds of thousands of \textcolor{red}{people} learned \textcolor{red}{how to do that}. \\
    \midrule
    Ground truth & And all at once \textcolor{blue}{I became a learner}. \\
    adapter-families & And all of a sudden, \textcolor{blue}{I became a learner}. \\
    adapter-fed & And all of a sudden, \textcolor{red}{it's happening}. \\
    \bottomrule
    \end{tabular}}
    \caption{Case study comparison of \emph{adapter-families} and \emph{adapter-fed}. Obvious translation mistakes are highlighted in \textcolor{red}{red} and corresponding correct translations are highlighted in \textcolor{blue}{blue}.}
    \label{tab:case_study}
\end{table}
\definecolor{+}{RGB}{0,150,0}

\begin{table}[h]
    \centering
    \resizebox{0.95\linewidth}{!}{ 
    \begin{tabular}{@{}ll|cc|cc@{}}
    
    \toprule
    & & \multicolumn{2}{c|}{TED2020} & \multicolumn{2}{c}{Europarl} \\
    \multicolumn{2}{c|}{Method} & Macro Avg. & Micro Avg. & Macro Avg. & Micro Avg.  \\
    \midrule
    \multirow{2}{*}{adapter-fed} & FedAvg & 34.31 & 31.70 & 22.20 & 23.00 \\
    & FedMean & \textcolor{+}{34.52} & \textcolor{+}{31.71} & \textcolor{+}{23.56} & \textcolor{+}{23.53} \\
    \midrule
    \multirow{2}{*}{adapter-random} & FedAvg & 34.79 & 32.16 & 25.19 & 25.76 \\
    & FedMean & \textcolor{+}{35.14} & \textcolor{+}{32.38} & \textcolor{+}{25.27} & \textcolor{red}{25.67} \\
    \midrule
    \multirow{2}{*}{adapter-gradients} & FedAvg & 34.46 & 31.96 & 25.64 & 25.84 \\
    & FedMean & \textcolor{+}{34.90} & \textcolor{+}{32.21} & \textcolor{red}{25.51} & \textcolor{red}{25.35} \\
    \midrule
    \multirow{2}{*}{adapter-families} & FedAvg & 34.83 & 32.09 & 25.47 & 26.02 \\
    & FedMean & \textcolor{+}{35.21} & \textcolor{+}{32.40} & \textcolor{+}{25.76} & \textcolor{+}{26.04} \\
    \bottomrule
    \end{tabular}}
    \caption{Comparison between FedAvg and FedMean (the aggregation method we use in Eq. \ref{eq:fedmean}). Results of FedMean are in \textcolor{+}{green} if better than those of FedAvg, otherwise in \textcolor{red}{red}. In most cases, FedMean yields performance gains.}
    \label{tab:improvement_analysis}
\end{table}

\section{Further Analysis}
\subsection{Case Study}
\label{subsec:case_study}
We select some representative cases of translations from method \emph{adapter-families} and \emph{adapter-shareAll}, which are shown in Table \ref{tab:case_study}, to further study the influence of our clustering strategies. We separate the mistakes into three categories. \par
\paragraph{Opposite semantics} In case 1 and case 2, \emph{adapter-fed} misses negation adverbs in predictions and results in totally opposite semantics.\par
\paragraph{Inaccurate words/phrases} In case 3, \emph{adapter-fed} translates adverbial of time into ``a day or two'', which should be ``four or five weeks'' actually. In case 4, \emph{adapter-fed} uses the expression ``save the world'', which differs from the original expression "change the world" in semantics.\par
\paragraph{Ambiguous semantics} In cases 5 and 6, \emph{adapter-fed} loses specific semantic information in ground truths. It fails to properly translate ``children'', ``science'' and ``become a learner'' and uses more ambiguous expressions instead.\par
In comparison, \emph{adapter-families} makes more accurate predictions in the above cases, which suggests that appropriate clustering strategies help the model produce better translations with improvements in semantics.

\begin{table*}[t]
    \centering
    \resizebox{0.95\textwidth}{!}{ 
    \begin{tabular}{@{}l|l|c|cccccccccccc|cc@{}}
    \toprule
    Method & adapters & Comm. Cost & de-fr & nl-pl & en-lt & fr-nl & it-sl & es-lv & pl-en  & sl-es & sl-lt & lt-de & lv-it & lv-pl & Macro Avg. & Micro Avg.  \\
    \midrule
    \multirow{4}{*}{\makecell[l]{Language families\\clustering}} & input-end & 2.7M & 30.21 & 19.04 & 29.98 & 20.39 & 14.93 & 28.09 & 41.64 & 32.08 & 21.39 & 21.62 & 19.89 & 20.73 & 25.00 & 25.12 \\
    & middle-layer & 2.7M & 30.12 & 18.88 & 29.84 & 21.27 & 15.15 & 28.16 & 41.33 & 31.92 & 21.06 & 21.68 & 20.99 & 20.74 & 25.10 & 25.30 \\
    & output-end & 2.7M & 30.30 & 19.37 & 30.33 & 22.17 & 16.06 & 27.96 & 41.39 & 31.76 & 20.03 & 21.75 & 21.97 & 21.15 & 25.35 & 25.64 \\
    & all adapters & 8.1M & 30.18 & 18.85 & 29.86 & 22.42 & 16.10 & 27.66 & 41.13 & 32.96 & 20.78 & 21.80 & 22.78 & 21.27 & 25.48 & 25.75 \\
    \bottomrule
    \end{tabular}}
    \caption{BLEU scores of different adapter pruning strategies. We acquire similar results with only 1/3 communication cost compared to keeping all adapter modules.}
    \label{tab:adapter_pruning}
\end{table*}

\subsection{Both FedMean and Clustering Contribute}
In our experiments, discrepancies in data come from two aspects: data quantity skew and linguistic discrepancy (language difference). We adjust the aggregation algorithm from Eq.~\eqref{eq:fedavg} to Eq.~\eqref{eq:fedmean}, which we call \emph{FedMean} here, to tackle the problem of quantity skew. Moreover, we propose clustering strategies to prevent clients from receiving conflicting parameters from dissimilar languages. To explore how these two methods contribute to improvement in performance, we further conduct experiments with the aggregation algorithm changed to FedAvg (see Eq.~\eqref{eq:fedavg}) and keep other training settings unchanged. 

As the results displayed in Table \ref{tab:improvement_analysis},  clustering strategies bring more significant improvements in performance on the Europarl corpus than the TED2020 corpus. Since experiments on the Europarl corpus consists of more different languages and are conducted in a more complicated m2m setting, the problem of linguistic discrepancy is more severe for Europarl.
For the TED2020 corpus, changing the aggregation algorithm from FedAvg to FedMean leads to more significant improvements for methods with clustering strategies compared to \emph{adapter-fed}. In contrast, for the Europarl corpus, \emph{adapter-fed} substantially benefits from FedMean, while FedMean hardly brings any benefits to methods with clustering strategies, even causing performance drop in some cases.\par
Based on these observations, we contend that both aggregation algorithms and clustering strategies contribute to performance gain by alleviating data discrepancies. The specific extent of improvement depends on the extent of data quantity skew and linguistic discrepancy.

\subsection{Further Cost Saving by Adapter Pruning}
On top of the adapter tuning approach, adapter pruning techniques~\citep{ruckle2020adapterdrop, pfeiffer2020adapterfusion, karimi2021compacter} further compress the number of parameters to be updated. 
To further reduce the communication cost, we conduct an exploratory attempt to prune parts of adapter modules in both the encoder and the decoder.
We still choose the mBART-50 model, with 12 layers in the encoder and the decoder separately, to conduct the experiments. Specifically, we evenly separate all adapter modules we add in mBART into three parts: input-end adapters (adapter modules in the first 4 layers of the encoder or the decoder), middle-layer adapters (adapter modules in layers 5 to 8 of the encoder or the decoder), output-end adapters (adapter modules in the last 4 layers of the encoder or the decoder). 
In each strategy, only one part of the adapter modules is kept, so the communication cost is saved by two-thirds. 
We use \emph{adapter-families} as the baseline and train all models with one random seed. The rest settings stay the same as those in previous main experiments. 
The results are shown in Table \ref{tab:adapter_pruning}.\par
It is encouraging that pruning adapters do not result in a sharp decrease in performance. We observe that keeping output-end adapters achieves the highest score among the three pruning strategies, which suggests that adapters in the top layers play more important roles. Overall, the results indicate that it is possible to further reduce communication costs and it is worthwhile to explore more elaborate pruning techniques in future work.

\section{Related Work}
\paragraph{Federated Learning} was first proposed by~\citet{mcmahan2017communication} as a decentralized training framework.
Due to its decentralized and private nature, FL shows great potential in actual applications.
Recently, there has been a surge in the NLP community to explore the application of federated learning in diverse NLP tasks, such as emojis prediciton~\citep{gandhi2022federated}, named entity recognition~\citep{ge2020fedner}, and machine translation~\citep{roosta2021communication,passban2022training,weller2022pretrained}, etc.
\citet{roosta2021communication} first applied FL to NMT tasks.
However, training language models in the FL setting brings huge communication overheads.
To solve this problem, researchers have proposed to only exchange some dedicated ``Controller''~\citep{roosta2021communication} layers between the server and clients.
Moreover, \citet{passban2022training} introduced parameter pruning strategies to reduce communication bandwidth.
Our methods with adapter modules have advantages in communication efficiency with fewer parameters to be transferred compared to Controller (see Appendix \ref{appendix:controller}), and other parameter pruning strategies can also be applied to our adapter modules to further reduce communication costs.\looseness=-1
\paragraph{Multilingual Neural Machine Translation} (MNMT) trains a single model to handle translation between multiple language pairs~\citep{johnson2017google,aharoni2019massively,zhang2020improving}.
Moreover, MNMT significantly reduces training and inference costs by eliminating the need to train models for each language pair.
Massively pre-trained multilingual models have been used for MNMT, such as mBART-50~\citep{Tang2020MultilingualTW} and M2M100~\citep{fan2021beyond}.
In recent years, adapter has become a popular method in MNMT~\citep{bapna-firat-2019-simple,cooper-stickland-etal-2021-recipes,philip-etal-2020-monolingual,ustun2021multilingual,chronopoulou2022language} due to its high parameter efficiency and transferability between tasks.\looseness=-1

Different from previous works on this topic, inspired by recent progress in improving the efficiency of NLP methods~\citep{strubell2019energy,li2021cascadebert,xu2021survey,li2021dynamic},
we focus on communication efficiency in FL-MNMT and make the first effort to introduce adapter modules in order to reduce communication costs. We also apply different clustering strategies to resolve the issue of conflicting parameters stemming from data discrepancy.

\section{Conclusion}
In this paper, we introduce adapter modules to PLMs for the Fed-MNMT problem to boost communication efficiency. We reduce the communication cost by over 98\% and make the training process of Fed-MNMT practical. To deal with the problem of performance drop after introducing adapter modules, we propose different clustering strategies to separate clients into different groups to avoid the negative influence of data discrepancy. We surpass the direct baseline with a substantial gap, especially in the more complicated multi-to-multi translation setting.\par
Furthermore, our analytic experiments indicate that both aggregation algorithms in server and clustering strategies affect the performance of Fed-MNMT. We also explore the possibility of further reducing communication costs by pruning adapter modules and find that adapters in top layers are more significant for translation performance. \par
In future work, we will explore more well-designed clustering strategies and attach other parameter-efficient techniques to adapter to further reduce the parameters to be transferred.

\section*{Limitations}
First, in this work, we assume that clustering in the encoder and the decoder is only related to the source and target languages, respectively. Actually, both parameters in the encoder and the decoder are influenced by source and target languages simultaneously. Therefore, our assumption may lead to a performance drop. In future work, we plan to explore more complicated clustering strategies. \par 
Moreover, our \emph{adapter-families} method depends on prior linguistic knowledge. Its actual effectiveness can be affected by the distribution of language families/groups in clients. Our methods mainly apply to comparably uniform language distribution. \par
In addition, the effectiveness of our methods on other PLMs needs to be verified. However, it is easy to transfer our methods to other models so it will not be a challenging problem.


\section*{Acknowledgements}
We thank all reviewers for their insightful comments and suggestions. This work is supported by Natural Science Foundation of China (NSFC) No.62176002. We sincerely thank Jingjing Xu for her valuable suggestions. Xu Sun is the corresponding author.

\bibliography{anthology,custom}

\begin{thebibliography}{35}
\expandafter\ifx\csname natexlab\endcsname\relax\def\natexlab#1{#1}\fi

\bibitem[{Aharoni et~al.(2019)Aharoni, Johnson, and
  Firat}]{aharoni2019massively}
Roee Aharoni, Melvin Johnson, and Orhan Firat. 2019.
\newblock \href {https://doi.org/10.18653/v1/N19-1388} {Massively multilingual
  neural machine translation}.
\newblock In \emph{Proceedings of the 2019 Conference of the North {A}merican
  Chapter of the Association for Computational Linguistics: Human Language
  Technologies, Volume 1 (Long and Short Papers)}, pages 3874--3884,
  Minneapolis, Minnesota. Association for Computational Linguistics.

\bibitem[{Bapna and Firat(2019)}]{bapna-firat-2019-simple}
Ankur Bapna and Orhan Firat. 2019.
\newblock \href {https://doi.org/10.18653/v1/D19-1165} {Simple, scalable
  adaptation for neural machine translation}.
\newblock In \emph{Proceedings of the 2019 Conference on Empirical Methods in
  Natural Language Processing and the 9th International Joint Conference on
  Natural Language Processing (EMNLP-IJCNLP)}, pages 1538--1548, Hong Kong,
  China. Association for Computational Linguistics.

\bibitem[{Chronopoulou et~al.(2022)Chronopoulou, Stojanovski, and
  Fraser}]{chronopoulou2022language}
Alexandra Chronopoulou, Dario Stojanovski, and Alexander Fraser. 2022.
\newblock \href {https://arxiv.org/abs/2209.15236} {Language-family adapters
  for multilingual neural machine translation}.
\newblock \emph{ArXiv preprint}, abs/2209.15236.

\bibitem[{Cooper~Stickland et~al.(2021)Cooper~Stickland, Li, and
  Ghazvininejad}]{cooper-stickland-etal-2021-recipes}
Asa Cooper~Stickland, Xian Li, and Marjan Ghazvininejad. 2021.
\newblock \href {https://doi.org/10.18653/v1/2021.eacl-main.301} {Recipes for
  adapting pre-trained monolingual and multilingual models to machine
  translation}.
\newblock In \emph{Proceedings of the 16th Conference of the European Chapter
  of the Association for Computational Linguistics: Main Volume}, pages
  3440--3453, Online. Association for Computational Linguistics.

\bibitem[{Fan et~al.(2020)Fan, Bhosale, Schwenk, Ma, El-Kishky, Goyal, Baines,
  Celebi, Wenzek, Chaudhary, Goyal, Birch, Liptchinsky, Edunov, Grave, Auli,
  and Joulin}]{fan2020englishcentric}
Angela Fan, Shruti Bhosale, Holger Schwenk, Zhiyi Ma, Ahmed El-Kishky,
  Siddharth Goyal, Mandeep Baines, Onur Celebi, Guillaume Wenzek, Vishrav
  Chaudhary, Naman Goyal, Tom Birch, Vitaliy Liptchinsky, Sergey Edunov,
  Edouard Grave, Michael Auli, and Armand Joulin. 2020.
\newblock \href {http://arxiv.org/abs/2010.11125} {Beyond english-centric
  multilingual machine translation}.

\bibitem[{Fan et~al.(2021)Fan, Bhosale, Schwenk, Ma, El-Kishky, Goyal, Baines,
  Celebi, Wenzek, Chaudhary et~al.}]{fan2021beyond}
Angela Fan, Shruti Bhosale, Holger Schwenk, Zhiyi Ma, Ahmed El-Kishky,
  Siddharth Goyal, Mandeep Baines, Onur Celebi, Guillaume Wenzek, Vishrav
  Chaudhary, et~al. 2021.
\newblock Beyond english-centric multilingual machine translation.
\newblock \emph{J. Mach. Learn. Res.}, 22(107):1--48.

\bibitem[{Firat et~al.(2016)Firat, Cho, and Bengio}]{firat2016multi}
Orhan Firat, Kyunghyun Cho, and Yoshua Bengio. 2016.
\newblock \href {https://doi.org/10.18653/v1/N16-1101} {Multi-way, multilingual
  neural machine translation with a shared attention mechanism}.
\newblock In \emph{Proceedings of the 2016 Conference of the North {A}merican
  Chapter of the Association for Computational Linguistics: Human Language
  Technologies}, pages 866--875, San Diego, California. Association for
  Computational Linguistics.

\bibitem[{Gandhi et~al.(2022)Gandhi, Mehta, Parekh, Waghela, D'Mello, and
  Talat}]{gandhi2022federated}
Deep Gandhi, Jash Mehta, Nirali Parekh, Karan Waghela, Lynette D'Mello, and
  Zeerak Talat. 2022.
\newblock \href {https://arxiv.org/abs/2211.06401} {A federated approach to
  predicting emojis in hindi tweets}.
\newblock \emph{ArXiv preprint}, abs/2211.06401.

\bibitem[{Ge et~al.(2020)Ge, Wu, Wu, Qi, Huang, and Xie}]{ge2020fedner}
Suyu Ge, Fangzhao Wu, Chuhan Wu, Tao Qi, Yongfeng Huang, and Xing Xie. 2020.
\newblock Fedner: Privacy-preserving medical named entity recognition with
  federated learning.
\newblock \emph{arXiv e-prints}, pages arXiv--2003.

\bibitem[{Houlsby et~al.(2019)Houlsby, Giurgiu, Jastrzebski, Morrone,
  de~Laroussilhe, Gesmundo, Attariyan, and Gelly}]{houlsby2019parameter}
Neil Houlsby, Andrei Giurgiu, Stanislaw Jastrzebski, Bruna Morrone, Quentin
  de~Laroussilhe, Andrea Gesmundo, Mona Attariyan, and Sylvain Gelly. 2019.
\newblock \href {http://proceedings.mlr.press/v97/houlsby19a.html}
  {Parameter-efficient transfer learning for {NLP}}.
\newblock In \emph{Proceedings of the 36th International Conference on Machine
  Learning, {ICML} 2019, 9-15 June 2019, Long Beach, California, {USA}},
  volume~97 of \emph{Proceedings of Machine Learning Research}, pages
  2790--2799. {PMLR}.

\bibitem[{Johnson et~al.(2017)Johnson, Schuster, Le, Krikun, Wu, Chen, Thorat,
  Vi{\'e}gas, Wattenberg, Corrado, Hughes, and Dean}]{johnson2017google}
Melvin Johnson, Mike Schuster, Quoc~V. Le, Maxim Krikun, Yonghui Wu, Zhifeng
  Chen, Nikhil Thorat, Fernanda Vi{\'e}gas, Martin Wattenberg, Greg Corrado,
  Macduff Hughes, and Jeffrey Dean. 2017.
\newblock \href {https://doi.org/10.1162/tacl_a_00065} {{G}oogle{'}s
  multilingual neural machine translation system: Enabling zero-shot
  translation}.
\newblock \emph{Transactions of the Association for Computational Linguistics},
  5:339--351.

\bibitem[{Karimi~Mahabadi et~al.(2021)Karimi~Mahabadi, Henderson, and
  Ruder}]{karimi2021compacter}
Rabeeh Karimi~Mahabadi, James Henderson, and Sebastian Ruder. 2021.
\newblock Compacter: Efficient low-rank hypercomplex adapter layers.
\newblock \emph{Advances in Neural Information Processing Systems},
  34:1022--1035.

\bibitem[{Koehn(2005)}]{koehn2005europarl}
Philipp Koehn. 2005.
\newblock \href {https://aclanthology.org/2005.mtsummit-papers.11} {{E}uroparl:
  A parallel corpus for statistical machine translation}.
\newblock In \emph{Proceedings of Machine Translation Summit X: Papers}, pages
  79--86, Phuket, Thailand.

\bibitem[{Li et~al.(2021{\natexlab{a}})Li, Lin, Chen, Ren, Li, Zhou, and
  Sun}]{li2021cascadebert}
Lei Li, Yankai Lin, Deli Chen, Shuhuai Ren, Peng Li, Jie Zhou, and Xu~Sun.
  2021{\natexlab{a}}.
\newblock Cascadebert: Accelerating inference of pre-trained language models
  via calibrated complete models cascade.
\newblock In \emph{Findings of the Association for Computational Linguistics:
  EMNLP 2021}, pages 475--486.

\bibitem[{Li et~al.(2021{\natexlab{b}})Li, Lin, Ren, Li, Zhou, and
  Sun}]{li2021dynamic}
Lei Li, Yankai Lin, Shuhuai Ren, Peng Li, Jie Zhou, and Xu~Sun.
  2021{\natexlab{b}}.
\newblock Dynamic knowledge distillation for pre-trained language models.
\newblock In \emph{Proceedings of the 2021 Conference on Empirical Methods in
  Natural Language Processing}, pages 379--389.

\bibitem[{McMahan et~al.(2017)McMahan, Moore, Ramage, Hampson, and
  y~Arcas}]{mcmahan2017communication}
Brendan McMahan, Eider Moore, Daniel Ramage, Seth Hampson, and
  Blaise~Ag{\"{u}}era y~Arcas. 2017.
\newblock \href {http://proceedings.mlr.press/v54/mcmahan17a.html}
  {Communication-efficient learning of deep networks from decentralized data}.
\newblock In \emph{Proceedings of the 20th International Conference on
  Artificial Intelligence and Statistics, {AISTATS} 2017, 20-22 April 2017,
  Fort Lauderdale, FL, {USA}}, volume~54 of \emph{Proceedings of Machine
  Learning Research}, pages 1273--1282. {PMLR}.

\bibitem[{Niu and Deng(2022)}]{niu2022federated}
Yifan Niu and Weihong Deng. 2022.
\newblock Federated learning for face recognition with gradient correction.
\newblock In \emph{Proceedings of the AAAI Conference on Artificial
  Intelligence}, volume~36, pages 1999--2007.

\bibitem[{Passban et~al.(2022)Passban, Roosta, Gupta, Chadha, and
  Chung}]{passban2022training}
Peyman Passban, Tanya Roosta, Rahul Gupta, Ankit Chadha, and Clement Chung.
  2022.
\newblock \href {https://doi.org/10.18653/v1/2022.naacl-main.186} {Training
  mixed-domain translation models via federated learning}.
\newblock In \emph{Proceedings of the 2022 Conference of the North American
  Chapter of the Association for Computational Linguistics: Human Language
  Technologies}, pages 2576--2586, Seattle, United States. Association for
  Computational Linguistics.

\bibitem[{Pfeiffer et~al.(2021)Pfeiffer, Kamath, R{\"u}ckl{\'e}, Cho, and
  Gurevych}]{pfeiffer2020adapterfusion}
Jonas Pfeiffer, Aishwarya Kamath, Andreas R{\"u}ckl{\'e}, Kyunghyun Cho, and
  Iryna Gurevych. 2021.
\newblock \href {https://doi.org/10.18653/v1/2021.eacl-main.39}
  {{A}dapter{F}usion: Non-destructive task composition for transfer learning}.
\newblock In \emph{Proceedings of the 16th Conference of the European Chapter
  of the Association for Computational Linguistics: Main Volume}, pages
  487--503, Online. Association for Computational Linguistics.

\bibitem[{Pfeiffer et~al.(2020)Pfeiffer, R{\"u}ckl{\'e}, Poth, Kamath,
  Vuli{\'c}, Ruder, Cho, and Gurevych}]{pfeiffer2020AdapterHub}
Jonas Pfeiffer, Andreas R{\"u}ckl{\'e}, Clifton Poth, Aishwarya Kamath, Ivan
  Vuli{\'c}, Sebastian Ruder, Kyunghyun Cho, and Iryna Gurevych. 2020.
\newblock \href {https://doi.org/10.18653/v1/2020.emnlp-demos.7}
  {{A}dapter{H}ub: A framework for adapting transformers}.
\newblock In \emph{Proceedings of the 2020 Conference on Empirical Methods in
  Natural Language Processing: System Demonstrations}, pages 46--54, Online.
  Association for Computational Linguistics.

\bibitem[{Philip et~al.(2020)Philip, Berard, Gall{\'e}, and
  Besacier}]{philip-etal-2020-monolingual}
Jerin Philip, Alexandre Berard, Matthias Gall{\'e}, and Laurent Besacier. 2020.
\newblock \href {https://doi.org/10.18653/v1/2020.emnlp-main.361} {Monolingual
  adapters for zero-shot neural machine translation}.
\newblock In \emph{Proceedings of the 2020 Conference on Empirical Methods in
  Natural Language Processing (EMNLP)}, pages 4465--4470, Online. Association
  for Computational Linguistics.

\bibitem[{Post(2018)}]{post-2018-call}
Matt Post. 2018.
\newblock \href {https://doi.org/10.18653/v1/W18-6319} {A call for clarity in
  reporting {BLEU} scores}.
\newblock In \emph{Proceedings of the Third Conference on Machine Translation:
  Research Papers}, pages 186--191, Brussels, Belgium. Association for
  Computational Linguistics.

\bibitem[{Rebuffi et~al.(2017)Rebuffi, Bilen, and
  Vedaldi}]{rebuffi2017learning}
Sylvestre{-}Alvise Rebuffi, Hakan Bilen, and Andrea Vedaldi. 2017.
\newblock \href
  {https://proceedings.neurips.cc/paper/2017/hash/e7b24b112a44fdd9ee93bdf998c6ca0e-Abstract.html}
  {Learning multiple visual domains with residual adapters}.
\newblock In \emph{Advances in Neural Information Processing Systems 30: Annual
  Conference on Neural Information Processing Systems 2017, December 4-9, 2017,
  Long Beach, CA, {USA}}, pages 506--516.

\bibitem[{Reimers and Gurevych(2020)}]{reimers-2020-multilingual-sentence-bert}
Nils Reimers and Iryna Gurevych. 2020.
\newblock \href {https://doi.org/10.18653/v1/2020.emnlp-main.365} {Making
  monolingual sentence embeddings multilingual using knowledge distillation}.
\newblock In \emph{Proceedings of the 2020 Conference on Empirical Methods in
  Natural Language Processing (EMNLP)}, pages 4512--4525, Online. Association
  for Computational Linguistics.

\bibitem[{Roosta et~al.(2021)Roosta, Passban, and
  Chadha}]{roosta2021communication}
Tanya Roosta, Peyman Passban, and Ankit Chadha. 2021.
\newblock \href {https://arxiv.org/abs/2112.06135} {Communication-efficient
  federated learning for neural machine translation}.
\newblock \emph{ArXiv preprint}, abs/2112.06135.

\bibitem[{R{\"u}ckl{\'e} et~al.(2021)R{\"u}ckl{\'e}, Geigle, Glockner, Beck,
  Pfeiffer, Reimers, and Gurevych}]{ruckle2020adapterdrop}
Andreas R{\"u}ckl{\'e}, Gregor Geigle, Max Glockner, Tilman Beck, Jonas
  Pfeiffer, Nils Reimers, and Iryna Gurevych. 2021.
\newblock \href {https://doi.org/10.18653/v1/2021.emnlp-main.626}
  {{AdapterDrop}: {O}n the efficiency of adapters in transformers}.
\newblock In \emph{Proceedings of the 2021 Conference on Empirical Methods in
  Natural Language Processing}, pages 7930--7946, Online and Punta Cana,
  Dominican Republic. Association for Computational Linguistics.

\bibitem[{Strubell et~al.(2019)Strubell, Ganesh, and
  McCallum}]{strubell2019energy}
Emma Strubell, Ananya Ganesh, and Andrew McCallum. 2019.
\newblock Energy and policy considerations for deep learning in nlp.
\newblock In \emph{Proceedings of the 57th Annual Meeting of the Association
  for Computational Linguistics}, pages 3645--3650.

\bibitem[{Tan et~al.(2019)Tan, Chen, He, Xia, Qin, and
  Liu}]{tan2019multilingual}
Xu~Tan, Jiale Chen, Di~He, Yingce Xia, Tao Qin, and Tie-Yan Liu. 2019.
\newblock \href {https://doi.org/10.18653/v1/D19-1089} {Multilingual neural
  machine translation with language clustering}.
\newblock In \emph{Proceedings of the 2019 Conference on Empirical Methods in
  Natural Language Processing and the 9th International Joint Conference on
  Natural Language Processing (EMNLP-IJCNLP)}, pages 963--973, Hong Kong,
  China. Association for Computational Linguistics.

\bibitem[{Tang et~al.(2020)Tang, Tran, Li, Chen, Goyal, Chaudhary, Gu, and
  Fan}]{Tang2020MultilingualTW}
Y.~Tang, C.~Tran, Xian Li, Peng-Jen Chen, Naman Goyal, Vishrav Chaudhary,
  Jiatao Gu, and Angela Fan. 2020.
\newblock Multilingual translation with extensible multilingual pretraining and
  finetuning.
\newblock \emph{ArXiv}, abs/2008.00401.

\bibitem[{{\"U}st{\"u}n et~al.(2021){\"U}st{\"u}n, Berard, Besacier, and
  Gall{\'e}}]{ustun2021multilingual}
Ahmet {\"U}st{\"u}n, Alexandre Berard, Laurent Besacier, and Matthias
  Gall{\'e}. 2021.
\newblock \href {https://doi.org/10.18653/v1/2021.emnlp-main.533} {Multilingual
  unsupervised neural machine translation with denoising adapters}.
\newblock In \emph{Proceedings of the 2021 Conference on Empirical Methods in
  Natural Language Processing}, pages 6650--6662, Online and Punta Cana,
  Dominican Republic. Association for Computational Linguistics.

\bibitem[{Wang et~al.(2020)Wang, Liu, Liang, Joshi, and
  Poor}]{wang2020tackling}
Jianyu Wang, Qinghua Liu, Hao Liang, Gauri Joshi, and H.~Vincent Poor. 2020.
\newblock \href
  {https://proceedings.neurips.cc/paper/2020/hash/564127c03caab942e503ee6f810f54fd-Abstract.html}
  {Tackling the objective inconsistency problem in heterogeneous federated
  optimization}.
\newblock In \emph{Advances in Neural Information Processing Systems 33: Annual
  Conference on Neural Information Processing Systems 2020, NeurIPS 2020,
  December 6-12, 2020, virtual}.

\bibitem[{Weller et~al.(2022)Weller, Marone, Braverman, Lawrie, and
  Van~Durme}]{weller2022pretrained}
Orion Weller, Marc Marone, Vladimir Braverman, Dawn Lawrie, and Benjamin
  Van~Durme. 2022.
\newblock \href {https://doi.org/10.18653/v1/2022.naacl-main.101} {Pretrained
  models for multilingual federated learning}.
\newblock In \emph{Proceedings of the 2022 Conference of the North American
  Chapter of the Association for Computational Linguistics: Human Language
  Technologies}, pages 1413--1421, Seattle, United States. Association for
  Computational Linguistics.

\bibitem[{Xu et~al.(2021)Xu, Zhou, Fu, Zhou, and Li}]{xu2021survey}
Jingjing Xu, Wangchunshu Zhou, Zhiyi Fu, Hao Zhou, and Lei Li. 2021.
\newblock A survey on green deep learning.
\newblock \emph{arXiv preprint arXiv:2111.05193}.

\bibitem[{Zhang et~al.(2020)Zhang, Williams, Titov, and
  Sennrich}]{zhang2020improving}
Biao Zhang, Philip Williams, Ivan Titov, and Rico Sennrich. 2020.
\newblock \href {https://doi.org/10.18653/v1/2020.acl-main.148} {Improving
  massively multilingual neural machine translation and zero-shot translation}.
\newblock In \emph{Proceedings of the 58th Annual Meeting of the Association
  for Computational Linguistics}, pages 1628--1639, Online. Association for
  Computational Linguistics.

\bibitem[{Zhao et~al.(2018)Zhao, Li, Lai, Suda, Civin, and
  Chandra}]{zhao2018federated}
Yue Zhao, Meng Li, Liangzhen Lai, Naveen Suda, Damon Civin, and Vikas Chandra.
  2018.
\newblock \href {https://arxiv.org/abs/1806.00582} {Federated learning with
  non-iid data}.
\newblock \emph{ArXiv preprint}, abs/1806.00582.

\end{thebibliography}
\bibliographystyle{acl_natbib}

\appendix
\section{Training Data Sizes}
The specific training data size of each language pair is shown in Table \ref{tab:data_size}.
\label{appendix:data_size}
\begin{table}[h]
    \centering
    \resizebox{0.95\linewidth}{!}{
    \begin{tabular}{@{}l|lcc@{}}
        \toprule
        Corpus & \makecell[l]{Source Language \\ Family/Group} & Language Pair & Dataset Size \\
        \midrule
        \multirow{8}{*}{TED2020} & \multirow{2}{*}{Sino-Tibetan} & zh->en & 9984 \\
        & \multirow{2}{*}{} & th->en & 4992 \\
        \cmidrule{2-4}
        & \multirow{2}{*}{Afro-asiatic} & ar->en & 9984 \\
        & \multirow{2}{*}{} & he->en & 1920 \\
        \cmidrule{2-4}
        & \multirow{2}{*}{Uralic} & fi->en & 1920 \\
        & \multirow{2}{*}{} & et->en & 1920 \\
        \cmidrule{2-4}
        & \multirow{2}{*}{Indo-European} & ru->en & 9984 \\
        & \multirow{2}{*}{} & sl->en & 1920 \\
        \midrule
        \midrule
        \multirow{12}{*}{Europarl} & \multirow{3}{*}{Germanic} & de->fr & 11648 \\
        & \multirow{3}{*}{} & nl->pl & 3584 \\
        & \multirow{3}{*}{} & en->lt & 3712 \\
        \cmidrule{2-4}
        & \multirow{3}{*}{Romance} & fr->nl & 12160 \\
        & \multirow{3}{*}{} & it->sl & 3456 \\
        & \multirow{3}{*}{} & es->lv & 3584 \\
        \cmidrule{2-4}
        & \multirow{3}{*}{Slavic} & pl->en & 3712 \\
        & \multirow{3}{*}{} & sl->es & 3584 \\
        & \multirow{3}{*}{} & sl->lt & 3584 \\
        \cmidrule{2-4}
        & \multirow{3}{*}{Baltic} & lt->de & 3328 \\
        & \multirow{3}{*}{} & lv->it & 3584 \\
        & \multirow{3}{*}{} & lv->pl & 3712 \\
    \bottomrule
    \end{tabular}}
    \caption{Detailed data sizes of language pairs from Europarl corpus.}
    \label{tab:data_size}
\end{table}
\section{Complete Results of Ablation Study}
We show the complete results on all language pairs of the ablation study in Table \ref{tab:ablation_study}.\par
We find that only applying clustering to the decoder acquires the highest scores on 7 out of the total 12 language pairs. To our surprise, \emph{adapter-families} method fails to reach the best performance in average scores, which has been explained in \S~\ref{subsec:ablation_study} in the main text.

\section{Uniform Data Distribution}
We also conduct experiments on TED2020 with each client owning training data of equal size. The results are shown in Table \ref{tab:uniform_results}. We observe that the performance gaps between different methods are similar to those in Table \ref{tab:ted_results}. Notably, \emph{Adapter-families} beats \emph{adapter-random} by a slight margin. Both clustering strategies acquire obvious performance improvement compared to the baseline \emph{adapter-fed}.
These empirical results verify that our methods apply to various data distributions.

\begin{table*}[t]
\centering
\resizebox{0.95\textwidth}{!}{ 
\begin{tabular}{@{}l|cccccccccccc|cc@{}}
\toprule
Method & de-fr & nl-pl & en-lt & fr-nl & it-sl & es-lv & pl-en  & sl-es & sl-lt & lt-de & lv-it & lv-pl & Macro Avg. & Micro Avg.  \\
\midrule
adapter-fed & 29.66 & 17.35 & 30.22 & 16.89 & 11.95 & 27.91 & 37.31 & 31.04 & 20.13 & 21.18 & 17.95 & 19.97 & 23.46 & 23.44 \\
\quad + encoder clustering & 30.50 & 19.24 & 29.85 & 22.94 & \textbf{18.56} & 28.01 & \textbf{41.93} & 33.25 & 20.84 & \textbf{22.09} & \textbf{23.34} & 20.82 & 25.95 & 26.19 \\
\quad + decoder clustering & \textbf{31.19} & \textbf{19.41} & \textbf{30.53} & \textbf{23.06} & 18.26 & \textbf{28.20} & 41.71 & \textbf{33.56} & \textbf{21.26} & 21.96 & 23.03 & 21.06 & \textbf{26.10} & \textbf{26.42} \\
adapter-families & 30.18 & 18.85 & 29.86 & 22.42 & 16.10 & 27.66 & 41.13 & 32.96 & 20.78 & 21.80 & 22.78 & \textbf{21.27} & 25.48 & 25.75 \\
\bottomrule
\end{tabular}}
\caption{BLEU scores of applying language families strategies to only the encoder and the decoder. \emph{Adapter-fed} corresponds to clustering in neither the encoder or the decoder. \emph{Adapter-families} denotes clustering in both the encoder and the decoder.}
\label{tab:ablation_study}
\end{table*}
\begin{table*}[t]
    \small
    \centering
    \resizebox{0.8\textwidth}{!}{
    \begin{tabular}{@{}l|cccccccc|c@{}}
    \toprule
   Method  & zh-en & th-en & ar-en & he-en & fi-en & et-en & ru-en & sl-en & Avg.\\
    \midrule
    model-fed & 25.58 & 22.90 & 39.63 & 46.12 & 34.78 & 34.41 & 30.42 & 51.32 & 35.64 \\
    \midrule
    adapter-fed & 24.83 & 16.79 & \textbf{40.02} & \textbf{45.99} & 33.55 & 33.67 & 29.61 & \textbf{52.20} & 34.58 \\
    adapter-random & \textbf{25.30} & \textbf{22.33} & 39.58 & 45.59 & 34.58 & 34.31 & 30.12 & 51.52 & 35.41 \\
    adapter-families & 25.04 & 22.08 & 39.50 & 45.74 & \textbf{34.91} & \textbf{34.65} & \textbf{30.16} & 51.38 & \textbf{35.43} \\
    \bottomrule
    \end{tabular}}
    \caption{BLEU scores on the TED2020 corpus with equivalent data size for each language pair. Best scores are highlighted in bold except \emph{model-fed}.}
    \label{tab:uniform_results}
\end{table*}

\section{Comparison to Controllers}
\label{appendix:controller}
Controllers \citep{roosta2021communication} only exchange 8 layers in a 32-layer Transformer (4 from encoder and 4 from decoder) between the server and clients, which means that they reduce the communication cost by approximately 66\% (the number of layers in the original model without Controllers is 24).
Compared with Controllers, we introduce adapter modules in Fed-MNMT without the need to define additional layers.
Besides, our methods transmit a much smaller amount of parameters in client-to-server exchanges than using Controllers.
Therefore, our proposal is superior to Controllers in terms of communication efficiency.

\end{document}